\documentclass[conference]{IEEEtran}
\IEEEoverridecommandlockouts
\input epsf
\usepackage{graphicx}
\usepackage{cite}
\usepackage{amssymb}
\usepackage{amsmath}
\usepackage{float}
\usepackage{booktabs}
\usepackage{tikz}
\usetikzlibrary{shapes,positioning}
\usepackage[margin=0.75in]{geometry}

\def\BibTeX{{\rm B\kern-.05em{\sc i\kern-.025em b}\kern-.08em
    T\kern-.1667em\lower.7ex\hbox{E}\kern-.125emX}}

\begin{document}

\makeatletter
\newcommand{\newlineauthors}{%
  \end{@IEEEauthorhalign}\hfill\mbox{}\par
  \mbox{}\hfill\begin{@IEEEauthorhalign}
}
\makeatother

\title{One-Class Classification for Intrusion Detection on Vehicular Networks\\}

\author{\IEEEauthorblockN{Jake Guidry\IEEEauthorrefmark{1},
Fahad Sohrab\IEEEauthorrefmark{3},
Raju Gottumukkala\IEEEauthorrefmark{1}
Satya Katragadda\IEEEauthorrefmark{2} and
Moncef Gabbouj\IEEEauthorrefmark{3}}
\IEEEauthorblockA{\IEEEauthorrefmark{1}Mechanical Engineering, University of Louisiana at Lafayette, United States}
\IEEEauthorblockA{\IEEEauthorrefmark{2}Informatics Research Institute, University of Louisiana at Lafayette, United States}
\IEEEauthorblockA{\IEEEauthorrefmark{3}Faculty of Information Technology and Communication Sciences, Tampere University, Finland}
jake.guidry1@lousiana.edu, fahad.sohrab@tuni.fi, raju.gottumukkala@louisiana.edu,\\ satyasivakumar@gmail.com, moncef.gabbouj@tuni.fi}

\maketitle

\begin{abstract}
Controller Area Network bus systems within vehicular networks are not equipped with the tools necessary to ward off and protect themselves from modern cyber-security threats. Work has been done on using machine learning methods to detect and report these attacks, but common methods are not robust towards unknown attacks. These methods usually rely on there being a sufficient representation of attack data, which may not be available due to there either not being enough data present to adequately represent its distribution or the distribution itself is too diverse in nature for there to be a sufficient representation of it. With the use of one-class classification methods, this issue can be mitigated as only normal data is required to train a model for the detection of anomalous instances. Research has been done on the efficacy of these methods, most notably One-Class Support Vector Machine and Support Vector Data Description, but many new extensions of these works have been proposed and have yet to be tested for injection attacks in vehicular networks. In this paper, we investigate the performance of various state-of-the-art one-class classification methods for detecting injection attacks on Controller Area Network bus traffic. We investigate the effectiveness of these techniques on attacks launched on Controller Area Network buses from two different vehicles during normal operation and while being attacked. We observe that the Subspace Support Vector Data Description method outperformed all other tested methods with a Gmean of about 85\%.
\end{abstract}
\IEEEoverridecommandlockouts
\IEEEpeerreviewmaketitle
 \section{Introduction} \label{sec:introduction}
Different kinds of cyber-security vulnerabilities can potentially render vehicles prone to several types of attacks, such as taking control of the vehicle, Denial of Service (DOS) attacks, and spoofing attacks \cite{lee2017otids}. As emerging technologies push devices across all domains to become more connected than ever before, the internal infrastructure of modern vehicles is still based upon antiquated standards and regulations from a time long before this interconnectivity was thought possible. Within the core of all vehicles lies the same internal communication protocol, the Controller Area Network (CAN) protocol, which was designed before the advent of WiFi and Bluetooth-enabled vehicles. The vehicles' CAN networks were designed for quick and efficient transmission of messages with no cyber-security features because they were not necessary for the time. But in a modern vehicle, this same protocol is used when it is exposed to external networks such as the Internet. These modern features open vehicles' networks to new vectors of attack that they were never designed for. This openness leaves vehicles as prime targets for well-known and widely used cyber-security attacks \cite{carsten2015vehicle,checkoway2011comprehensive,studnia2013survey}. The gravity of these threats is exhibited by researchers who performed hacks on Tesla vehicles that disabled critical safety features \cite{lambert2019hackers}. As these threats become more apparent, automobile manufacturers are implementing countermeasures such as data encryption and message authentication in order to mitigate the likelihood of a successful attack. These additions would make newer models of vehicles more resilient to these attacks, but vehicles that were manufactured before these changes would still have no protection against these threats.

Traditional machine learning techniques have been employed for the purpose of detecting injection-based cyber-security attacks. These techniques read the traffic of CAN messages on the vehicle's CAN bus in order to make a decision about whether the traffic is normal or anomalous. Often, several different characteristics of the bus's traffic are extracted, such as the inter-signal arrival time, the frequency of messages on the bus, and appearances of message sequences \cite{moore2017modeling, taylor2015frequency, marchetti2017anomaly}. However, these techniques rely on training data containing samples of both normal and anomalous data, meaning that prior knowledge of the specific type of attack is required in order to detect it in the future. These models and techniques perform well in scenarios where the models were given sufficient training data of the type of attack used, but their performance is affected when the models are tasked with detecting attacks for which they were given little to no training data. This poses an issue where in order for these models to remain updated and secure against new and emerging threats, they must be retrained with new data of these attacks. Research has been done on trying to mitigate this issue of detecting anomalies consisting of "unknown" attacks. In 2016, researchers Bezemskij et al. proposed a knowledge-based approach to detect anomalies on a robotic vehicle \cite{bezemskij2016behaviour}. The researchers attempted to improve the detection of unknown attacks by adjusting weights that were trained on known attacks to accentuate the features of the data that indicated anomalous activity. The researchers reported that this technique proved effective in detecting unknown attacks within their experimental setup, but this technique may prove ineffective against unknown attacks whose features do not follow the trend of features of known attacks. Researchers in \cite{song2021self} proposed another self-supervised technique to detect unknown attacks. The researchers create a pseudo-normal data generator to generate data that mimics normal operating data on a CAN network but lies just outside of the normal data within a given feature space. A binary classification model is then trained with the noised data labeled as anomalous data to make a decision boundary between the normal data and noised data. For these techniques to be useful, either data of these attacks must be available, or a method for generating data that mimics attack data must be available.

We propose the use of one-class classification methods for the purpose of detecting anomalous data on a vehicle's CAN bus. One-class classification methods have been used successfully for identifying data that falls outside the bounds of normal or expected behavior across multiple domains, such as the medical field with detecting myocardial infractions \cite{degerli2022early} and taxa identification with identifying rare benthic macroinvertebrates \cite{sohrab2020boosting}. These approaches use one-class classification for identifying data outside the norm using features from their respective datasets. Our proposed approach follows this paradigm, where we only train the model on data collected from the CAN bus under normal behavior. This methodology differs from already established techniques of intrusion detection on CAN buses by only requiring normal data rather than training a model on both normal and anomalous data. The main contribution of this work is that because the model will be trained without anomalous data from any specific type of cyber-attack, we claim that this approach will be more suited to detecting unknown attacks within a vehicular network.

The structure of this paper is as follows: Section \ref{sec:relatedwork} provides an overview of work associated with intrusion detection in the context of cyber-security on CAN buses as well as the application of one-class classification techniques for the purpose of intrusion detection. Section \ref{sec:methodology} provides the methodologies used for the various one-class classification techniques that were applied to our dataset. Section \ref{sec:experimentalsetup} shows our experimental setup, which includes the physical layout of our data collection apparatus, the feature generation, and the methods used for generating anomalous data for testing purposes. Section \ref{sec:conclusion} then presents the conclusions that were made from our experiments as well as potential improvements and future work.
\section{Related Work} \label{sec:relatedwork}
The vulnerability of the CAN bus system has been discussed extensively in the literature. Researchers Koscher et al. provide a good overview of the vulnerabilities associated with the CAN bus system \cite{koscher2010experimental}. Researchers describe several key pitfalls with the system's security, such as the broadcast nature of the system, the vulnerability of the system to DoS attacks, the system's lack of authentication, and weak protection against gaining access to ECU's control mechanisms. The paper presents a multitude of various actions that they were able to exploit on the vehicle. In 2015, researchers Miller and Valasek exploited the vulnerabilities of the CAN bus system on a Jeep Cherokee and were able remotely to execute these attacks through the vehicles' cellular interfaces. The researchers were able to perform actions such as disabling critical systems, such as the vehicle's braking system, and manipulating the steering wheel.

Intrusion detection systems (IDS) are software packages or hardware devices that monitor the flow of information across a CAN bus and identify if the CAN bus is secure or compromised. These IDSs can be broadly categorized into four categories with a few exceptions that do not fit these descriptions: fingerprint-based methods, parameters monitoring-based methods, information-theoretic-based methods, and machine learning-based methods \cite{wu2019survey}. Our methods fall into the category of machine learning-based, so this section will discuss the work done within this domain.

Different types of machine learning approaches have been used to detect intrusions. Most of these techniques utilize models that rely on data from both normal and anomalous data. These methods include LSTM neural networks, deep neural networks, and hidden Markov models. \cite{taylor2016anomaly, kang2016intrusion, narayanan2015using}. These approaches yield good results, typically higher than 90\% accuracy for intrusion detection, but they rely on a good description of anomalous data. Another issue with some of these approaches is the computational complexity, especially with the deep learning options. For our proposed intrusion detection framework, the intrusion detection model needs to have a low enough computational complexity to be real-time, but for some of these deep learning methods, a high amount of computation is necessary for the level of performance required to be effective. The drawbacks of these established approaches are the requirement of sufficient anomalous data for training the models, the lack of effectiveness against unknown attacks, and the potential computational cost.

Techniques have been developed for the purpose of remedying similar challenges. One-class classification techniques are useful for scenarios where classification is required, but there is a severe under-representation of one of the classes, referred to as the non-target class \cite{sohrab2023graph}. These techniques train their models exclusively on a single class, the target class, in order to provide a classification of whether a sample is part of the target class or not part of the target class. Various one-class methods have been used across multiple domains for detecting faults or anomalies with regard to the normal behavior of an operation. In 2005, researchers Shin et al. applied OC-SVMs for machine fault detection based on vibration measurements \cite{shin2005one}. The researchers claim that the application of an OC-SVM can reduce the cost of maintenance and operation by more accurately detecting system faults. A paper by Sanchez-Hernandez et al. reports that with the use of Support Vector Data Description (SVDD), a one-class classification technique, they were able to achieve accuracies higher than conventional multi-class techniques when classifying a single target class \cite{sanchez2007one}.

There has also been an application of these models specifically geared toward cyber-security intrusion detection. In \cite{li2003improving}, researchers tested one-class models on a dataset that simulated TCP/IP traffic over a local-area network. This paper claimed that the one-class-based SVM outperformed other standard classification techniques such as clustering, K-Nearest Neighbors, Na{\"i}ve Bayes, and standard SVM. Researchers from the Georgia Institute of Technology also combined OC-SVMs with other one-class classifiers for ensemble intrusion detection systems monitoring HTTP (Hypertext Transfer Protocol) requests \cite{perdisci2006using}. Along with these applications toward more internet-structured networks, work has been done within the domain of vehicular networks specifically. In \cite{theissler2014anomaly}, OC-SVM and SVDD methods are used to detect anomalies on vehicular networks. This paper focuses on detecting faults within the vehicle's infrastructure rather than cyber-security attacks but still uses one-class models for intrusion detection based on data extracted from CAN messages. Although, these methods are also being used as intrusion detection systems on CAN networks, as seen in papers by Maglaras and Taylor et al., where OC-SVM techniques are used to detect intrusions on vehicular networks \cite{maglaras2015novel, taylor2015frequency}.
\section{Methodology} \label{sec:methodology}

Our approach is meant to fit into a multi-layered process where our methods act on a security module that monitors the CAN bus traffic on a vehicle and reports to a centralized authority or management system. We are presenting a potential workflow on which our methods can be implemented for the purpose of intrusion detection on a vehicle's CAN bus. Fig. 1 shows the process that our implementation will fit into.

\begin{figure}[]
    \label{fig:tikzworkflow}
    \begin{center}
        \begin{tikzpicture}[font=\small,thick,node distance=.5cm]
            
            \node[draw,
                minimum width=2.5cm,
                minimum height=.75cm
                ] (block1) {Vehicle Operation};
            
            \node[draw,
                align=center,
                below=of block1,
                minimum width=2.5cm,
                minimum height=.75cm
                ] (block2) {Data Collection};
                
            \node[draw,
                align=center,
                below=of block2,
                minimum width=2.5cm,
                minimum height=.75cm
                ] (block3) {Feature Extraction};
                
            \node[draw,
                align=center,
                below=of block3,
                minimum width=2.5cm,
                minimum height=.75cm
                ] (block4) {One-Class Classification};
                
            \node[draw,
                align=center,
                below=of block4,
                minimum width=2.5cm,
                minimum height=.75cm
                ] (block5) {Reporting};
            
            \draw[-latex] (block1) edge (block2)
                (block2) edge (block3)
                (block3) edge (block4)
                (block4) edge (block5);
    
        \end{tikzpicture}
    \end{center}
    \caption{Workflow for intrusion detection process}
\end{figure}
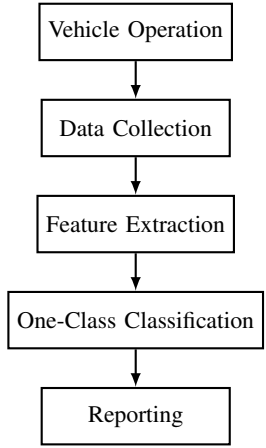

\subsection{Vehicle Operation}

The process begins with the normal use of a vehicle. While a vehicle is being used, there will be CAN bus traffic being transmitted from one electronic control unit (ECU) to another. These ECUs require messages to be transmitted and received very quickly and efficiently, which is the main criterion that the CAN protocol was built for. Within any given second, the can bus can be populated with dozens of CAN messages, each with its own CAN ID and data payload.

\subsection{Data Collection}

During vehicle operation, traffic will be monitored and recorded by some security module that has access to the CAN bus of the vehicle. Once this data has been collected for a certain timeframe, feature extraction will be performed on the raw CAN messages that were collected during the said timeframe. These messages contain the timestamp, ID, and data payload of each message. The process in which the data is collected and the hardware and software used are explained in more detail in section \ref{sec:experimentalsetup}.

\subsection{Feature Extraction}

Once the raw CAN messages were recorded, three different features were constructed from each unique CAN ID, each of which is shown below.

\vspace{0.25cm}
\begin{itemize}
    \item \textbf{Average Frequency of Appearance of a CAN ID}: How frequently and given CAN ID appears on the CAN bus
    \begin{equation}
        f = \sum_{k=2} ^{j} {\frac{j-1}{t_{i,k+1}-t_{i,k}}}
        \label{eqn:frequency}
    \end{equation}
    
    \item \textbf{Average Time Interval Between Consecutive Appearance of a CAN ID}: Time delta change between an appearance of any given CAN ID and its next appearance on the bus
    \begin{equation}
        \Delta t = \sum_{k=2} ^{j} {\frac{t_{i,k+1}-t_{i,k}}{j-1}}
        \label{eqn:timeDiff}
    \end{equation}
    
    \item \textbf{Standard Deviation of Transmission Times of CAN IDs}: Standard Deviation of a selected CAN ID's transmission times across a set period of time.
    \begin{equation}
        s = \sqrt{\frac{1}{j} \sum_{k=2} ^{j} {(t_{i,k}-\overline{t_{i}})^2}}
        \label{eqn:stdev}
    \end{equation}

\end{itemize}

where $f$ is the frequency, $\Delta t$ is the average time interval, $s$ is the standard deviation, $i$ is the index of the CAN ID, $j$ is the number of CAN messages associated with a specific ID, and $t$ is the timestamp of an ID's appearance.

This produced 3*i number features where i is the number of unique IDs present in normal CAN bus traffic. The number of IDs present will change between vehicles and manufacturers, but the process used for feature extraction is the same for any given CAN bus. These features were chosen as they do not rely on the data itself within each message but rather on the characteristics of how these messages appear. This was chosen so that these methods can be applied to any vehicular can bus. Each ECU within a vehicle is expected to transmit messages at expected intervals, so these features are very consistent under normal conditions \cite{merchant2019}.

\subsection{One-Class Classification} \label{sec:anomalydetection}

The next step involves feeding these features into a one-class classification model. Following is a list of the tested models and a brief description of them.

\begin{itemize}
    \item \textbf{Support Vector Data Description (SVDD)}\cite{tax2004support}: This method takes training data from the target class only to fit a spherically shaped boundary between normal data and outliers (anomalous data). To account for the possibility of outliers within the training set, slack variables are introduced to allow data samples of the target class to extend beyond the defined boundary of the discriminatory hyper-sphere while penalizing large distances between the sample and the boundary. This results in the following equation being minimized:
    
    \begin{equation}
        F(R,\textbf{a}) = R^2 + C\sum_{i} {\xi_i}
    \end{equation}
    
    with the constraints that most samples are within the hyper-sphere:
    
    \begin{equation}
        ||\textbf{x}_i-\textbf{a}||^2 \leq R^2 + \xi_i, ~~ \xi_i \geq 0 ~~ \forall i
    \end{equation}
    
    where $F$ is the function to minimize, $R$ is the radius of the hyper-sphere, $\textbf{a}$ is the center of the hyper-sphere, $C$ is a hyperparameter used to control the trade-off between the hypersphere's volume and the errors, and $\xi$ is the slack variable for each sample. Applying Lagrangian optimization yields the following equation to optimize:
    
    \begin{equation}
        \begin{split}
            & L(R,\textbf{a},\alpha_i,\gamma_i,\xi_i) = \\
            & R^2 + C\sum_i{\xi_i}-\sum_i{\gamma_i \xi_i} \\ 
            -&\sum_i{\alpha_i\left\{R^2+\xi_i-(\textbf{x}_i^\intercal\textbf{x}_i-2\textbf{a}^\intercal\textbf{x}_i+\textbf{a}^\intercal\textbf{a})\right\}}
        \end{split}
    \end{equation}
    
    with the Lagrange multipliers $\alpha_i \geq 0$ and $\gamma_i \geq 0$.
    
    \item \textbf{Subspace Support Vector Data Description (S-SVDD)}\cite{sohrab2018subspace}: This method builds upon SVDD by transforming the data from the given feature space to an optimized lower-dimensional feature space. Along with training a model to fit a hyper-sphere to the data, the model is iteratively trained to determine a transformation matrix $\textbf{Q}$ to optimally reduce the dimension of the feature space from $D$ to $d$ where $\textbf{Q} \in \mathbb{R}^{d \times D}$ such that
    
    \begin{equation}
        \textbf{y}_i = \textbf{Qx}_i, ~~ i = 1,...,N
    \end{equation}
    
    where $\left\{\textbf{x}_i\right\}, ~~ i = 1,...,N$ is the original training set in dimension $D$ and $\left\{\textbf{y}_i\right\}, ~~ i = 1,...,N$ is the transformed training set in dimension $d$. 
    Applying Lagrangian optimization to this new method yields the following equation to optimize:
    
    \begin{equation}
        \begin{split}
            & L(R,\textbf{a},\alpha_i,\gamma_i,\xi_i, \textbf{Q}) = \\
            & R^2 + C\sum_i{\xi_i}-\sum_i{\gamma_i \xi_i} \\ 
            -&\sum_i{\alpha_i(R^2+\xi_i-\textbf{x}^\intercal_i\textbf{Q}^\intercal_i\textbf{Q}\textbf{x}_i+2\textbf{a}^\intercal\textbf{Q}\textbf{x}_i-\textbf{a}^\intercal\textbf{a})}
        \end{split}
    \end{equation}

    \item \textbf{Ellipsoidal Support Vector Data Description (E-SVDD)}: This method builds upon the SVDD by fitting an ellipsoidal boundary to the data rather than a spherical boundary. Fitting a hyperellipsoid to the data rather than a hypersphere provides a greater level of generalizability by allowing the discriminate boundary more degrees of freedom. In \cite{sohrab2020ellipsoidal}, finding an optimized subspace for ellipsoidal data description has been proposed.
    \item \textbf{Graph-Embedded Support Vector Data Description (GE-SVDD)}: This method continues the work of SVDD by implementing graph embedding. This technique is suitable for datasets where the data can be represented in a graph structure. This is most applicable for problems where each sample contains data as well as information representing a relationship between the data.
    
    \item \textbf{One-Class Support Vector Machine (OC-SVM)}\cite{scholkopf1999support}: This technique closely resembles that of a regular support vector machine but is altered to fit the paradigm of a one-class model. Rather than circumscribing a hypersphere about the target class, this method discriminates the data by means of a hyperplane, as is also the case with a typical SVM model. Rather than using this hyperplane to discriminate between two different classes, this model fits the hyperplane to a single target class while maximizing the distance from the hyperplane to the origin of the feature space. In other words, the model is fit to discriminate the target class from the origin while fitting the hyperplane as far from the origin as possible.
    
    \item \textbf{One-Class Graph-Embedded Support Vector Machine (OC-GE-SVM)}\cite{mygdalis2016graph}: This method follows the same motivation as GE-SVDD where graph embedding is implemented into OC-SVM.
\end{itemize}

\subsection{Reporting}

Once a decision is made during the intrusion detection phase, the resulting information is relayed wirelessly from the security module on the vehicle to some central authority regarding the status of the vehicle. This central authority could be some building or energy management system that has the ability to authenticate or block a vehicle's access to some vehicle-to-vehicle network. This management system would have the power and responsibility to respond to a cyber-security threat by denying network access to infected vehicles and quarantining them off from other non-infected vehicles in an effort to impede the propagation of malware.


\section{Experimental Setup} \label{sec:experimentalsetup}

The used data set consisted of regular and anomalous intra-vehicular communications on two different electric vehicles: a 2011 Nissan Leaf and Chevy Volt. The data collection process was identical for both vehicles, but the set of IDs between the vehicles is different because they are manufacturer-specific. Data was collected from the Primary and Electric Vehicle(EV) CAN buses on the vehicles, but the methods used for feature extraction and intrusion detection are not specific to any CAN bus. In order to access the CAN bus of these vehicles, an On-board Diagnostic (OBD) breakout box was connected to the OBD-II port located underneath the steering wheel of the vehicles. This enabled easy access to each of the individual pins on the OBD-II port. A CAN interface device was then used to exchange data between the vehicle's CAN bus and our computer's USB interface. Using the python-can python library on a Linux machine with socketCAN, a python script was created to collect and record all CAN traffic on a vehicle's CAN bus. Fig. 2 shows the setup and connections between components. 

\begin{figure}[b]
    \centering
    \includegraphics[width=.7\linewidth]{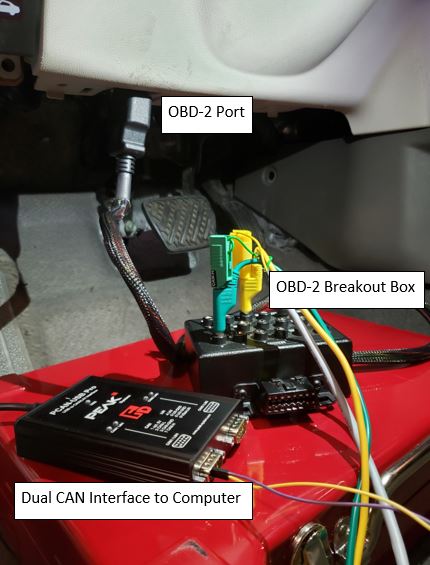}
    \caption{Wiring of CAN Bus Breakout Box to Nissan Leaf's OBD-II Port to computer}
    \label{fig:OBDbreakout}
\end{figure}

Normal data was collected from the vehicles as the vehicles were performing typical activities such as driving, parking, etc. Anomalous data was collected during the CyberAuto event, where various CAN bus injection attacks were conducted on the vehicles. These attacks include Random ID Attack, Zero ID Attack, and Replay Attack. These attacks were conducted using the same hardware setup used to collect the CAN data. While the data was collected, these attacks were deployed on the vehicle's primary and EV CAN buses. Shown below is a list and a  brief description of each of the attacks launched on the vehicles.

\begin{itemize}
    \item \textbf{Random ID Attack}: The Random ID Attack is a denial of service attack that floods the CAN bus of the vehicle with packets populated with random data that belong to random CAN IDs. The IDs used can be known or foreign.
    
    \item \textbf{Zero ID Attack}: The Zero ID Attack is a denial of service attack that floods the CAN bus of the vehicle with packets that belong to the CAN ID 0. The data payload for these packets can be empty or populated with data.
    
    \item \textbf{Replay Attack}: The Replay Attack is a denial of service attack that captures CAN packets from the CAN bus and replays them back onto the bus. This attack can replay single CAN messages or a sequence of CAN messages.
\end{itemize}

For each of the data collection sessions, all of the CAN traffic on the CAN bus was captured and saved into CSV files. The raw data saved into these CSV files contained each CAN message transmitted on the CAN bus during the data collection session. Each CAN message contained the CAN ID, the data payload, and the timestep of the message. In total, 1554 data samples were used to train each model, with 70\% used for training and 30\% used for testing.

\section{Results} \label{sec:results}

We discuss the results of the study by presenting the Gmean scores of each method described in section \ref{sec:methodology}, using the dataset collected as discussed in section \ref{sec:experimentalsetup}. We ran the models on each dataset, where the target class was the normal data. Consequently, each model was exclusively trained on the normal data. Subsequently, we tested the models on the entire test set, which includes both normal and anomalous data.

The Gmean score is a valuable metric for evaluating a model's performance when dealing with significant class imbalances, a common scenario in one-class classification methods. The Gmean score can be computed as $Gmean = \sqrt{TPR*TNR}$, where $TPR$ represents the true positive rate and $TNR$ represents the true negative rate. Our analysis includes tables of results that present performance metrics on the entire dataset, as well as augmented versions of the dataset. In the augmented versions, only a subset of the anomalous data was used to test the model. The purpose of this augmentation was to assess how effectively the model could detect individual types of attacks, particularly because certain cyber-attacks can be relatively straightforward to detect.

The reported results encompass all of the methods described in section \ref{sec:anomalydetection}, covering both their linear and nonlinear variants, where applicable. Furthermore, the study includes evaluations of four different variants of the S-SVDD method, identified in the tables as $\psi0$, $\psi1$, $\psi2$, or $\psi3$. The S-SVDD methodology employs regularization to determine which samples of the data are utilized to describe the class variance in the lower-dimensional optimized subspace.

\begin{table}[H]
    \centering
    \caption{Leaf Electric Vehicle}
    \label{tab:leafEV}
    \begin{tabular}{lllll}
        \hline
        \multicolumn{1}{|l|}{Target Class} & Normal & Random & Replay & \multicolumn{1}{l|}{Zero} \\ \hline
        \multicolumn{1}{|l|}{Linear}       &        &        &        &                           \\ \cline{1-1}
        SVDD                               & \textbf{0.84} & 0.78 & 0.65 & \textbf{1.00}                    \\
        S-SVDD-$\psi0$                           & \textbf{0.84} & 0.79 & 0.91 & 0.98                    \\
        S-SVDD-$\psi1$                           & \textbf{0.84} & 0.79 & \textbf{0.99} & 0.98                    \\
        S-SVDD-$\psi2$                           & \textbf{0.84} & 0.79 & 0.92 & 0.98                    \\
        S-SVDD-$\psi3$                           & \textbf{0.84} & 0.79 & 0.97 & 0.98                    \\
        E-SVDD                             & 0.83 & 0.70 & 0.58 & 0.96                    \\
        GE-SVDD                            & \textbf{0.84} & 0.82 & 0.94 & 0.86                    \\
        OC-SVM                             & 0.60 & 0.53 & 0.29 & 0.00                    \\
        GE-OC-SVM                          & 0.82 & 0.86 & 0.93 & 0.94                    \\ \cline{1-1}
        \multicolumn{1}{|l|}{Non-Linear}   &        &        &        &                           \\ \cline{1-1}
        SVDD                               & \textbf{0.84} & 0.79 & 0.65 & \textbf{1.00}                    \\
        S-SVDD-$\psi0$                           & \textbf{0.84} & 0.83 & 0.77 & 0.96                    \\
        S-SVDD-$\psi1$                           & \textbf{0.84} & 0.85 & 0.95 & 0.95                    \\
        S-SVDD-$\psi2$                           & \textbf{0.84} & \textbf{0.89} & 0.92 & 0.96                    \\
        S-SVDD-$\psi3$                           & \textbf{0.84} & 0.76 & 0.83 & 0.96                    \\
        E-SVDD                             & \textbf{0.84} & 0.80 & 0.67 & 0.89                    \\
        OC-SVM                             & 0.61 & 0.30 & 0.45 & 0.81                   
    \end{tabular}
\end{table}
\begin{table}[H]
    \centering
    \caption{Leaf Primary}
    \label{tab:leafPrimary}
    \begin{tabular}{lllll}
        \hline
        \multicolumn{1}{|l|}{Target Class} & Normal & Random & Replay & \multicolumn{1}{l|}{Zero} \\ \hline
        \multicolumn{1}{|l|}{Linear}       &        &        &        &                           \\ \cline{1-1}
        SVDD                               & \textbf{0.85} & 0.79 & 0.64 & 0.93                    \\
        S-SVDD-$\psi0$                           & 0.84 & 0.79 & 0.98 & 0.96                    \\
        S-SVDD-$\psi1$                           & 0.84 & 0.79 & \textbf{0.99} & 0.96                    \\
        S-SVDD-$\psi2$                           & 0.84 & 0.80 & 0.98 & 0.96                    \\
        S-SVDD-$\psi3$                           & 0.84 & 0.79 & 0.98 & 0.96                    \\
        E-SVDD                             & 0.83 & 0.76 & 0.58 & 0.93                    \\
        GE-SVDD                            & 0.83 & 0.81 & 0.92 & 0.90                    \\
        OC-SVM                             & 0.35 & 0.48 & 0.54 & 0.48                    \\
        GE-OC-SVM                          & 0.81 & \textbf{0.87} & 0.92 & 0.96                    \\ \cline{1-1}
        \multicolumn{1}{|l|}{Non-Linear}   &        &        &        &                           \\ \cline{1-1}
        SVDD                               & \textbf{0.85} & 0.79 & 0.64 & \textbf{1.00}                     \\
        S-SVDD-$\psi0$                           & 0.84 & 0.86 & 0.79 & 0.98                    \\
        S-SVDD-$\psi1$                           & 0.84 & 0.84 & 0.94 & 0.98                    \\
        S-SVDD-$\psi2$                           & 0.84 & \textbf{0.87} & 0.93 & 0.98                    \\
        S-SVDD-$\psi3$                           & 0.84 & 0.86 & 0.73 & 0.98                    \\
        E-SVDD                             & \textbf{0.85} & 0.85 & 0.65 & 0.93                    \\
        OC-SVM                             & 0.24 & 0.34 & 0.60 & 0.81                   
    \end{tabular}
\end{table}
\begin{table}[H]
    \centering
    \caption{Volt Electric Vehicle}
    \label{tab:voltEV}
    \begin{tabular}{lllll}
        \hline
        \multicolumn{1}{|l|}{Target Class} & Normal & Random & Replay & \multicolumn{1}{l|}{Zero} \\ \hline
        \multicolumn{1}{|l|}{Linear}       &        &        &        &                           \\ \cline{1-1}
        SVDD                               & 0.82 & 0.78 & 0.73 & 0.92                    \\
        S-SVDD-$\psi0$                           & 0.83 & 0.80 & 0.95 & 0.93                    \\
        S-SVDD-$\psi1$                           & 0.83 & 0.79 & \textbf{0.99} & 0.93                    \\
        S-SVDD-$\psi2$                           & 0.83 & 0.80 & 0.98 & 0.93                    \\
        S-SVDD-$\psi3$                           & 0.83 & 0.79 & 0.98 & 0.93                    \\
        E-SVDD                             & 0.81 & 0.76 & 0.57 & 0.92                    \\
        GE-SVDD                            & 0.83 & 0.83 & 0.92 & 0.94                    \\
        OC-SVM                             & 0.26 & 0.44 & 0.26 & 0.00                    \\
        GE-OC-SVM                          & 0.80 & 0.84 & 0.93 & 0.95                   \\ \cline{1-1}
        \multicolumn{1}{|l|}{Non-Linear}   &        &       &        &                           \\ \cline{1-1}
        SVDD                               & 0.82 & 0.78 & 0.65 & 0.96                    \\
        S-SVDD-$\psi0$                           & \textbf{0.84} & \textbf{0.86} & 0.83 & 0.95                    \\
        S-SVDD-$\psi1$                           & 0.70 & 0.82 & 0.91 & \textbf{0.98}                    \\
        S-SVDD-$\psi2$                           & 0.82 & 0.79 & 0.93 & \textbf{0.98}                    \\
        S-SVDD-$\psi3$                           & 0.69 & 0.77 & 0.82 & 0.94                    \\
        E-SVDD                             & 0.83 & 0.85 & 0.58 & 0.92                    \\
        OC-SVM                             & 0.64 & 0.30 & 0.60 & 0.80                   
    \end{tabular}
\end{table}
\begin{table}[H]
    \label{tab:voltPrimary}
    \centering
    \caption{Volt Primary}
    \begin{tabular}{lllll}
        \hline
        \multicolumn{1}{|l|}{Target Class} & Normal & Random & Replay & \multicolumn{1}{l|}{Zero} \\ \hline
        \multicolumn{1}{|l|}{Linear}       &        &        &        &                           \\ \cline{1-1}
        SVDD                               & 0.84 & 0.78 & 0.72 & \textbf{1.00}                    \\
        S-SVDD-$\psi0$                           & 0.84 & 0.79 & 0.97 & 0.95                    \\
        S-SVDD-$\psi1$                           & 0.84 & 0.79 & \textbf{0.99} & 0.95                    \\
        S-SVDD-$\psi2$                           & 0.84 & 0.78 & 0.98 & 0.95                    \\
        S-SVDD-$\psi3$                           & 0.84 & 0.79 & \textbf{0.99} & 0.95                    \\
        E-SVDD                             & 0.82 & 0.74 & 0.58 & \textbf{1.00}                    \\
        GE-SVDD                            & 0.84 & 0.81 & 0.88 & 0.94                    \\
        OC-SVM                             & 0.41 & 0.39 & 0.45 & 0.53                    \\
        GE-OC-SVM                          & 0.82 & \textbf{0.85} & 0.93 & 0.96                    \\ \cline{1-1}
        \multicolumn{1}{|l|}{Non-Linear}   &        &        &        &                           \\ \cline{1-1}
        SVDD                               & 0.84 & 0.80 & 0.81 & \textbf{1.00}                    \\
        S-SVDD-$\psi0$                           & 0.84 & 0.84 & 0.94 & \textbf{1.00}                    \\
        S-SVDD-$\psi1$                           & 0.84 & 0.79 & 0.94 & 0.97                    \\
        S-SVDD-$\psi2$                           & \textbf{0.98} & 0.78 & 0.96 & 0.96                    \\
        S-SVDD-$\psi3$                           & 0.84 & 0.77 & 0.84 & \textbf{1.00}                    \\
        E-SVDD                             & 0.84 & 0.81 & 0.65 & 0.97                    \\
        OC-SVM                             & 0.22 & 0.49 & 0.72 & 0.72                   
    \end{tabular}
\end{table}
The models used were able to generate a decision in real-time when using a window of one second to generate features. Tables I-IV show that most of the tested models were able to achieve a Gmean score of at least 80\%. The best-performing model for each category is highlighted in the tables. In all of our observed cases, the worst-performing method was OC-SVM, which is one of the most widely used one-class methods used in anomaly and intrusion detection systems. These experiments show that the newer adaptations of SVDD, S-SVDD especially, outperform OC-SVM when detecting anomalies in our dataset.

\section{Conclusion and Future Work} \label{sec:conclusion}
In this paper, we propose a simple framework for an anomaly detection system for CAN bus systems. For all four CAN buses, the one-class methods, especially S-SVDD, perform well in detecting cyber-security attacks on the CAN bus. We've shown that these techniques are suitable for anomaly detection systems on vehicular networks by virtue of the performance of these one-class models.

There is still room for improvement in regard to the application of these techniques for anomaly detection on vehicular networks. The features used in training the models were simple and trivial and may not have given a good description of the data to optimally train the models. A possible avenue for this discussion is the use of deep features or multimodal data for data description \cite{sohrab2021multimodal}. Also, because of how the proposed method extraction is structured, a predefined timeframe is required, which dictates the amount of time that CAN messages are collected before feature extraction. Depending on the timeframe used, this could lead to vulnerabilities with low-frequency attacks. This could potentially be solved by, instead of using a timeframe, considering the traffic of CAN data as a data stream.
\bibliography{refs}
\bibliographystyle{ieeetr}
\end{document}